\pdfoutput=1
\documentclass[11pt]{article}

\usepackage{acl}

\usepackage{times}
\usepackage{latexsym}
\usepackage{amsmath}
\usepackage{amsfonts}
\usepackage{booktabs}
\usepackage{multirow}
\usepackage{amsmath}
\usepackage{amssymb}
\usepackage{array}
\usepackage{xspace}
\usepackage{latexsym}
\usepackage{graphicx}
\usepackage{dblfloatfix}
\usepackage{xcolor}
\definecolor{darkgreen}{HTML}{006400}

\usepackage[T1]{fontenc}
\usepackage[utf8]{inputenc}

\usepackage{microtype}

\newcommand{\factgraph}{{\textsc{FactGraph}}\xspace}
\newcommand{\factgraphe}{{\textsc{FactGraph-E}}\xspace}
\newcommand{\factgraphtitle}{{\textsc{FactGraph}}\xspace}
\newcommand{\factcc}{{\textsc{FactCC}}\xspace}
\newcommand{\smatch}{{\textsc{Smatch-AMR}}}

\title{\factgraphtitle: Evaluating Factuality in Summarization \\
with Semantic Graph Representations}

\author{
\begin{minipage}[t]{\textwidth}
\centering
Leonardo F. R. Ribeiro$^{\dag}$\thanks{\hspace{0.2cm}Work done as an intern at Amazon Alexa AI.} \hspace{0.1cm}, Mengwen Liu$^{\ddag}$, Iryna Gurevych$^{\dag}$, \\
Markus Dreyer$^{\ddag}$, Mohit Bansal$^{\ddag,\S}$ \vspace{1mm} 
\end{minipage}
\\
\rule{0pt}{2.5ex}
  $^{\dag}$UKP Lab, Technical University of Darmstadt\\
  $^{\ddag}$Amazon Alexa AI, $^{\S}$UNC Chapel Hill \\
  {\small \texttt{ribeiro@aiphes.tu-darmstadt.de}, \texttt{\{mengwliu, mddreyer, mobansal\}@amazon.com}} \\
 {\small \texttt{gurevych@ukp.informatik.tu-darmstadt.de}, \texttt{mbansal@cs.unc.edu}}
}

\begin{document}
\maketitle
\begin{abstract}
Despite recent improvements in abstractive summarization, most current approaches generate summaries that are not \mbox{\emph{factually consistent}} with the source document, severely restricting their trust and usage in real-world applications. Recent works have shown promising improvements in factuality error identification using text or dependency arc entailments; however, they do not consider the entire semantic graph simultaneously. To this end, we propose \factgraph, a method that decomposes the document and the summary into structured \emph{meaning representations} (MR), which are more suitable for factuality evaluation. MRs describe core semantic concepts and their relations, aggregating the main content in both document and summary in a canonical form, and reducing data sparsity. \factgraph encodes such graphs using a graph encoder augmented with structure-aware adapters to capture interactions among the concepts based on the graph connectivity, along with text representations using an adapter-based text encoder. Experiments on different benchmarks for evaluating factuality show that \factgraph outperforms previous approaches by up to 15\%. Furthermore, \factgraph improves performance on identifying content verifiability errors and better captures subsentence-level factual inconsistencies.\footnote{Our code is publicly available at \url{https://github.com/amazon-research/fact-graph}}
\end{abstract}

\section{Introduction}

Recent summarization approaches based on pretrained language models (LM) have established a new level of performance~\cite{pmlr-v119-zhang20ae, lewis2020bart}, generating summaries that are grammatically fluent and capable of combining salient parts of the source document. However, current models suffer from a severe limitation, generating summaries that are \emph{not factually consistent}, that is, the content of the summary does not meet the facts of the source document, an issue also known as \emph{hallucination}. Previous studies~\cite{Cao_Wei_Li_Li_2018, falke-etal-2019-ranking, maynez-etal-2020-faithfulness, dreyer2021analyzing} report rates of hallucinations in generated summaries ranging from 30\% to over 70\%. In the face of such a challenge, recent works employ promising ideas such as question answering (QA)~\cite{durmus-etal-2020-feqa,nan-etal-2021-improving} and weakly supervised approaches~\cite{kryscinski-etal-2020-evaluating} to assess factuality. Another line of work explores dependency arc entailment to improve the localization of subsentence-level errors within generated summaries~\cite{goyal-durrett-2020-evaluating}.

 \begin{figure}[t]
    \centering
    \includegraphics[width=0.43\textwidth]{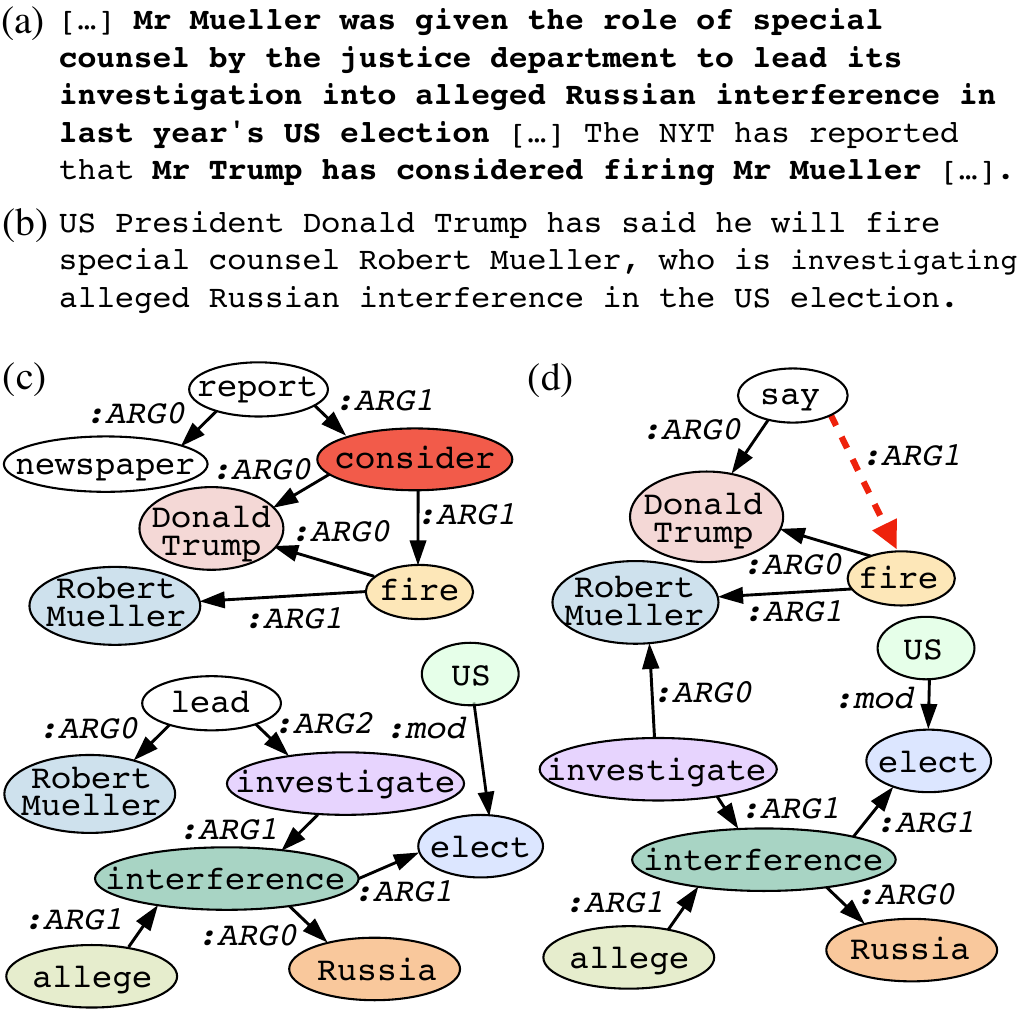}
    \caption{Example of (a) a document, (b) a summary, and (c) the corresponding document and (d) summary graph-based meaning representations. The summary graph does not contain the "consider" node, indicating a factual error (red dashed edge).
    } 
    \label{fig:example}
    \vspace{-3mm}
\end{figure}

However, these methods have a reduced correlation with human judgments and may not capture well semantic errors~\cite{pagnoni-etal-2021-understanding}. One reason for the poor performance is the lack of good quality factuality training data. Second, it is challenging to properly encode core semantic content from the document and summary~\cite{lee2021analysis} and reason over salient pieces of information in order to assess the summary factuality. Third, previous work ({\textsc{DAE}},~\citeauthor{goyal-durrett-2021-annotating},~\citeyear{goyal-durrett-2021-annotating}) treats semantic relations as isolated units, not simultaneously considering the entire semantic structure of \emph{both} document and summary texts.

To mitigate the above issues, we explore \emph{meaning representations} (MR) as a form of content representation for factuality evaluation. We present \factgraph, a novel graph-enhanced approach that incorporates core information from the document and the summary into the factuality model using graph-based MRs, which are more suitable for factuality evaluation: As shown in Figure~\ref{fig:example}, graph-based MRs capture semantic relations between entities, abstracting away from syntactic structure and producing a canonical representation of meaning. 

\begin{figure*}[t]
    \centering
    \includegraphics[width=0.85\textwidth]{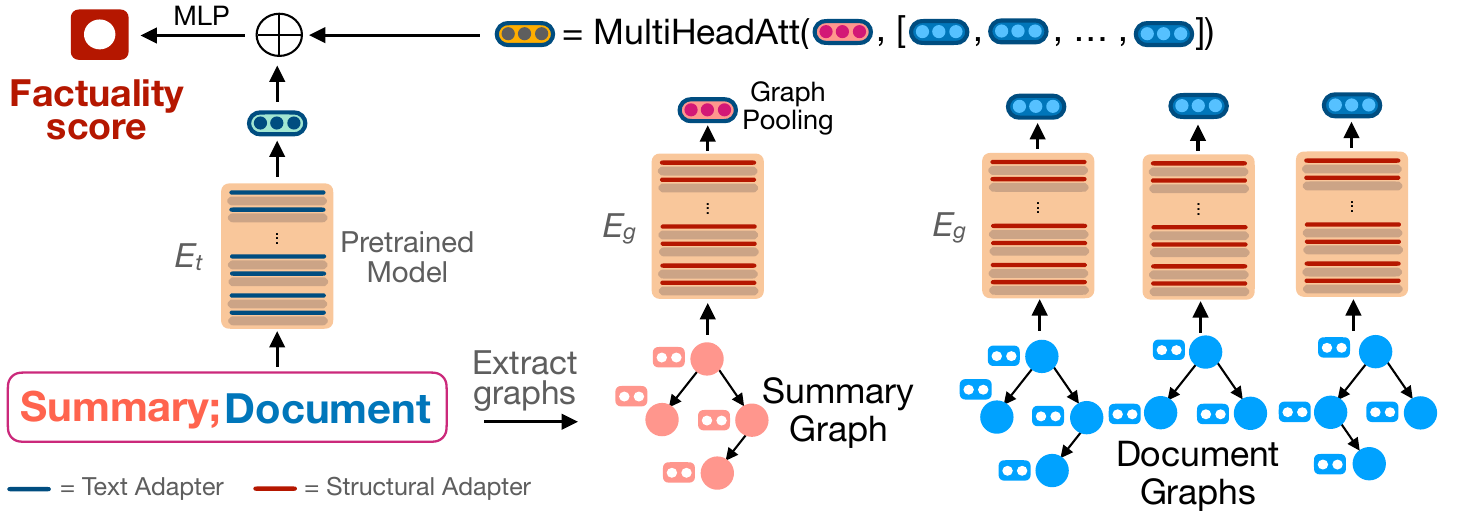}
    \caption{Overview of \factgraph. A sentence-level summary and document graphs are encoded by the graph encoder with structure-aware adapters. Text and graph encoders use the same pretrained model and only the adapters parameters are trained.}
    \label{fig:graphs}
    \vspace{-3mm}
\end{figure*}

Different from previous methods~\cite{kryscinski-etal-2020-evaluating,goyal-durrett-2021-annotating}, \factgraph is a dual approach which encodes both text and graph modalities, better integrating linguistic knowledge and structured semantic knowledge. As shown in Figure~\ref{fig:graphs}, it is composed of parameter-efficient text and graph encoders which share the same pretrained model and differ by their adapter weights~\cite{pmlr-v97-houlsby19a}. The texts from the document and summary are encoded using the adapter-based text encoder whereas the entire semantic structures that represent document and summary facts are used as input to the graph encoder augmented structure-aware adapters~\cite{ribeiro2021structural}. The representations of the two modalities thus are combined to generate the factuality score. 

In particular, we explore \emph{Abstract Meaning Representation} (AMR)~\cite{banarescu-etal-2013-abstract}, a semantic formalism that has received much research interest \cite{song-etal-2018-graph, doi:10.116200269, ribeiro-etal-2019-enhancing,ribeiro-etal-2021-smelting, opitz-etal-2020-amr, bamboo, fu-etal-2021-end} and has been shown to benefit downstream tasks such as spoken language understanding~\cite{damonte-etal-2019-practical}, machine translation \cite{doi:10.116200252}, commonsense reasoning~\cite{lim-etal-2020-know}, and question answering~\cite{kapanipathi-etal-2021-leveraging,bornea2021learning}.  

Intuitively, AMR provides important benefits: First, it encodes core concepts as it strives for a more logical and less syntactic representation, which has been shown to benefit text summarization~\cite{hardy-vlachos-2018-guided,dohare-etal-2018-unsupervised,lee2021analysis}. Furthermore, AMR captures semantics at a high level of abstraction explicitly modeling relations in the text and reducing the negative influence of diverse text surface variances with the same meaning. Lastly, recent studies~\cite{dreyer2021analyzing,ladhak2021faithful} demonstrate that there is a trade-off between factuality and abstractiveness. Structured semantic representations are potentially beneficial for reducing data sparsity and localizing generation errors in abstractive scenarios. Figure~\ref{fig:example} shows examples of (c) document and (d) summary AMRs, where the summary AMR is missing a crucial modifying node present in the document AMR, which indicates a factual error in the summary.

We consolidate a factuality dataset with human annotations derived from previous works ~\cite{wang-etal-2020-asking, kryscinski-etal-2020-evaluating,maynez-etal-2020-faithfulness,pagnoni-etal-2021-understanding}. This dataset is constructed from the widely-used CNN/DM~\cite{10.5555/2969239.2969428} and XSum~\cite{nallapati-etal-2016-abstractive2} benchmarks. Extensive experimental results demonstrate that \factgraph achieves substantial improvements over previous approaches, improving factuality performance by up to 15\% and correlation with human judgments by up to 10\%, capturing more content verifiability errors and better classifying factuality in semantic relations. 

\section{Related Work}
\paragraph{Evaluating Factuality.} Recently, there has been a surge of new methods for factuality evaluation in text generation, especially for summarization. \citet{falke-etal-2019-ranking} propose to rerank summary hypotheses generated via beam search based on entailment scores to the source document. \citet{kryscinski-etal-2020-evaluating} introduce \factcc, a model-based approach trained on artificially generated data, to measure if the summary can be entailed by the source document in order to assess the summary factuality. QA-based methods~\cite{wang-etal-2020-asking,durmus-etal-2020-feqa,DBLP:journals/corr/abs-2104-08202,nan-etal-2021-improving} generate questions from the document and summary, and compare the corresponding answers in order to assess factuality. \citet{xie2021factual} formulate causal relationships among the document, summary, and language prior to evaluate the factuality via counterfactual estimation. 

\paragraph{Categorizing Factual Errors.}A thread of analysis work has focused on identifying different categories of factual errors in summarization. \citet{maynez-etal-2020-faithfulness} show that semantic inference-based automatic measures are better representations of summarization quality, whereas \citet{pagnoni-etal-2021-understanding} propose a linguistically grounded typology of factual errors and develop a fine-grained benchmark for factuality evaluation, moving to a fine-grained measure, instead of using a binary evaluation. \citet{10.1162/tacl_a_00373} introduce different resources for summarization evaluation which include a toolkit for evaluating  summarization models.

\paragraph{Factuality versus Abstractiveness.} Recent works~\cite{dreyer2021analyzing,ladhak2021faithful} investigate the trade-off between factuality and abstractiveness of summaries and observe that factuality tends to drop with increased abstractiveness. Semantic graphs are uniquely suitable to detect factual errors in abstractive summaries as they abstract away from the lexical surface forms of documents and summaries, enabling direct comparisons of the underlying semantic concepts and relations of a document-summary pair.
\vspace{-1mm}

\paragraph{Graph-based Representations for Summarization.} A growing body of work focuses on using graph-based representations for improving summarization. Whereas different approaches encode graphs into neural models for multi-document summarization~\cite{fan-etal-2019-using,li-etal-2020-leveraging-graph, pasunuru-etal-2021-efficiently,wu-etal-2021-bass, chen-etal-2021-sgsum}, AMR structures have been shown to benefit both document representation and summary generation~\cite{liu-etal-2015-toward,liao-etal-2018-abstract,hardy-vlachos-2018-guided,dohare-etal-2018-unsupervised} and have the potential of improving controllability in summarization~\cite{lee2021analysis}. The above works are related to \factgraph as they use semantic graphs for content representation, but also different because they utilize graphs for the downstream summarization task, whereas \factgraph employs them for factuality evaluation.

\paragraph{Semantic Representations for Factuality Evaluation.} More closely related to our work, ~\citet{10.1145/3292500.3330955} extract tuples from the document and summary and measure the factual consistency by overlapping metrics. \citet{xu-etal-2020-fact} weight facts present in the source document according to the facts from the gold summary using contextual embeddings, and verify whether a generated summary is able to capture the same facts as the target.  Recently, dependency arc entailment ({\textsc{DAE}},~\citeauthor{goyal-durrett-2020-evaluating}, \citeyear{goyal-durrett-2020-evaluating}) is used to measure subsentence-level factuality by classifying pairs of words defined by dependency arcs which often describe semantic relations. However, \factgraph is considerably different from those approaches, since it explicitly encodes the entire graph semantic structure into the model. Moreover, while {\textsc{DAE}} considers semantic edge relations of the summary only, \factgraph encodes the semantic structures of both the input document and summary leading to better factuality performance at both sentence and subsentence levels. 

\section{\factgraphtitle Model}

We introduce \factgraph, a method that employs semantic graph representations for factuality evaluation in text summarization, describing its intuition (\S\ref{sec:intuition}) and defining it formally (\S\ref{sec:method}).

\subsection{Problem Statement}

Given a source document $D$ and a \emph{sentence-level} summary $S$, we aim to check whether $S$ is \emph{factual} with respect to $D$. For each sentence $d \in D$ we extract a semantic graph $\mathcal{G}_{d}$. Similarly, for the summary sentence $S$ we extract its semantic graph $\mathcal{G}_{s}$. We use texts and graphs from both document and summary for factuality evaluation. Sentence-level summary predictions can be aggregated to generate a factuality score for a multi-sentence summary. 

\subsection{Extracting Semantic Graphs}

We select AMR as our MR, but \factgraph can be used with other graph-based semantic representations, such as OpenIE~\cite{10.5555/1625275.1625705}. AMR is a linguistically-grounded semantic formalism that represents the meaning of a sentence as a rooted graph, where nodes are \emph{concepts} and edges are \emph{semantic relations}. AMR abstracts away from surface text, aiming to produce a more language-neutral representation of meaning. We use a state-of-the-art AMR parser~\cite{Micheleamr} to extract an AMR graph $\mathcal{G}_{a} = (\mathcal{V}_{a}, \mathcal{E}_{a}, \mathcal{R}_{a})$  with a node set $\mathcal{V}_{a}$ and labeled edges $ (u, r, v) \in \mathcal{E}_{a}$, where $u, v \in \mathcal{V}_{a}$ and $r \in \mathcal{R}_{a}$ is a relation type. Each $\mathcal{G}_{a}$ aims to explicitly represent the core concepts in each sentence. Figure~\ref{fig:example} shows an example of a (b) sentence and its (d) corresponding AMR graph.\footnote{Appendix \ref{appe:examples_amrs} presents other examples of AMRs extracted from sentences of documents and generated summaries.}

\paragraph{Graph Representation.}
We convert each $\mathcal{G}_{a}$ into a bipartite graph $\mathcal{G}_b = (\mathcal{V}_b, \mathcal{E}_b)$, replacing each labeled edge $(u,r,v) \in \mathcal{E}_{a}$ with two unlabeled edges $\{(u, r), (r, v)\} \in \mathcal{E}_{b}$. Similar to \citet{beck-etal-2018-graph}, this procedure transforms the graph into its unlabeled version. Pretrained models typically use a vocabulary with subword tokens, which makes it complicated to properly represent a graph using subword tokens as nodes. Inspired by~\citet{ribeiro2020modeling, ribeiro2021structural}, we transform each $\mathcal{G}_b$ into a new token graph $\mathcal{G} = (\mathcal{V}, \mathcal{E})$, where each token of a node $v_b \in \mathcal{V}_b$ becomes a node $v \in \mathcal{V}$. We convert each edge $ (u_b, v_b) \in \mathcal{E}_b$ into a set of edges and connect every token of $u_b$ to every token of $v_b$.

\subsection{Intuition of Semantic Representation}
\label{sec:intuition}

In order to represent facts to better assess the summary factuality, we draw inspiration from traditional approaches to summarization that condense the source document to a set of ``semantic units''~\cite{liu-etal-2015-toward, liao-etal-2018-abstract}. Intuitively, the semantic graphs from the source document represent the core factual information, explicitly modeling relations in the text, whereas the semantic summary graph captures the essential content information in a summary~\cite{lee2021analysis}. The document graphs can be compared with the summary graph,  measuring the degree of semantic overlap to assess factuality~\cite{cai-knight-2013-smatch}.

Recently, sets of fact triples from summaries were used to estimate factual accuracy~\cite{10.1145/3292500.3330955}. That approach is related to \factgraph as it uses graph-based MRs, but also different because it compares the reference and the generated summary, whereas we compare the generated summary with the input document. Moreover, differently from ~\citet{10.1145/3292500.3330955}, \factgraph explicitly encodes the semantic structures using a graph encoder and employs AMR as a semantic representation. Finally, in contrast to {\textsc{DAE}}~\cite{goyal-durrett-2021-annotating}, which focuses only on extracting summary graph representations, \factgraph uses semantic graphs for both document and summary.

\subsection{Model}
\label{sec:method}
Figure~\ref{fig:graphs} illustrates \factgraph, which is composed of text and graph encoders. The text encoder, denoted by $E_t$, uses a pretrained encoder $E$, augmented with adapter modules which receives the summary $S$ and document $D$ and outputs a contextual text representation. Conversely, the graph encoder, denoted by $E_g$, uses the same $E$, but is augmented with structure-aware adapters. $E_g$ receives the summary and multiple document semantic graphs corresponding to its sentences, and outputs graph-aware contextual representations that are used to generate the final graph representation. During training, only adapter weights are trained, whereas the weights from $E$ are kept frozen. Finally, both graph and text representations are concatenated and fed to a final classifier, which predicts whether the summary is factual or not.
\paragraph{Text Encoder.} We employ an adapter module before and after the feed-forward sub-layer of each layer of the encoder. We modify the adapter architecture from \citet{pmlr-v97-houlsby19a}. We compute the adapter representation for each token $i$ at each layer $l$, given the token representation $\boldsymbol{h}_{i}^{l}$, as follows:
\begin{align}
\hat{\boldsymbol{z}}^{l}_i &= \boldsymbol{W}_o^{l} ( \sigma (\boldsymbol{W}^{l}_p \,  {\scriptstyle\mathsf{LN}}(\boldsymbol{h}^{l}_i))) + \boldsymbol{h}^{l}_i \, ,
\end{align}
where $\sigma$ is the activation function and ${\scriptstyle\mathsf{LN}}(\cdot)$ denotes layer normalization. $\boldsymbol{W}^{l}_o \in \mathbb{R}^{d \times m}$ and $\boldsymbol{W}^{l}_p \in \mathbb{R}^{m \times d}$ are adapter parameters. The representation of the {\textsc{[CLS]}} token is used as the final textual representation, denoted by $\boldsymbol{t}$.

\paragraph{Graph Encoder.} In order to re-purpose the pretrained encoder to structured inputs, we employ a structural adapter~\cite{ribeiro2021structural}. In particular, for each node $v \in \mathcal{V}$, given the hidden representation $\boldsymbol{h}^{l}_v$, the encoder layer $l$ computes:
\begin{align}
\boldsymbol{g}^{l}_{v} &= {\mathsf{GraphConv}}_{l} ({\scriptstyle\mathsf{LN}}(\boldsymbol{h}^{l}_v),\!\{ {\scriptstyle\mathsf{LN}}(\boldsymbol{h}^{l}_u)\!: u \in {\scriptstyle\mathcal{N}(v)} \})\hspace{-.5em}  \nonumber \\
\boldsymbol{z}^{l}_v &= \boldsymbol{W}^{l}_{e} \sigma (\boldsymbol{g}^{l}_{v}) + \boldsymbol{h}^{l}_v \, ,
\end{align}
where $\mathcal{N}(v)$ is the neighborhood of the node $v$ in $\mathcal{G}$ and $\mathbf{W}^{l}_e \in \mathbb{R}^{d \times m}$ is an adapter parameter. $\mathsf{GraphConv}_{l}(\cdot)$ is the graph convolution that computes the representation of $v$ based on its \emph{neighbors} in the graph. We employ a Relational Graph Convolutional Network~\cite{Schlichtkrull2018ModelingRD} as graph convolution, which considers differences in the incoming and outgoing relations. Since AMRs are directed graphs, encoding edge directions is beneficial for downstream performance~\cite{ribeiro-etal-2019-enhancing}. The structural adapter is placed before, whereas the normal adapter is kept after the feed-forward sub-layer of each encoder layer.

We calculate the final representation of each graph from the pooling denoted as $\boldsymbol{z}^{\text{G}} = \{\mathbf{z}^{(L)}_v \mid v \in \mathcal{V} \}$, where $\mathbf{z}^{(L)}_v$ is the final representation of $v$. Thus, we use a multi-head self-attention~\cite{NIPS2017_7181} layer to estimate to what extent each sentence graph contributes to the document semantic representation based on the summary graph. This mechanism allows encoding a global document representation based on the summary graph. In particular, each attention head computes:
\begin{equation}
	\begin{split}
		\alpha_{i} &= \texttt{Attn}(\boldsymbol{z}^{\text{G}}_{s}, \boldsymbol{z}^{\text{G}}_{i}), \\
		\boldsymbol{g} &= \sum\nolimits_{i=1}^k  \alpha_{i} \, \boldsymbol{W}_{r} \, \boldsymbol{z}^{\text{G}}_{i}, \\
	\end{split}
\end{equation}
where $\boldsymbol{z}^{\text{G}}_{s}$ is the final representation of $\mathcal{G}_s$, $k$ is the number of considered sentence graphs from the input document and $\mathbf{W}_r \in \mathbb{R}^{d \times d}$ is a parameter.

The final representation is derived from the text and graph representations, $\boldsymbol{q} = [\boldsymbol{t}; \boldsymbol{g}]$, and fed into a classification layer that outputs a probability distribution over the labels $y$ = \{Factual, Non-Factual\}.

\subsection{Edge-level Factuality Model}
\label{sec:method-edges}
Inspired by~\citet{goyal-durrett-2021-annotating}, we evaluate the factuality at the edge level. In this setup, we use the same text and graph encoders; however, we encode the semantic graphs differently. In particular, we concatenate $\mathcal{G}_{s}$ with each $\mathcal{G}_{d} \in D$ and feed the concatenation to the graph encoder. The representation of a node $v \in \mathcal{G}_{s}$ is calculated as:  
\begin{equation}
	\label{eq:edgerep}
	\begin{split}
		\boldsymbol{r}_{t_v} &= \sum\nolimits_{w \in A(v)} E_t(D; S)_{w} \\
		\boldsymbol{r}_{g_v} &= \sum\nolimits_{d=1}^k E_g(\mathcal{G}_{s}; \mathcal{G}_{d})_{v} \\
		\boldsymbol{r}_v &= [\boldsymbol{r}_{t_v} ; \boldsymbol{r}_{g_v}] \\
	\end{split}
\end{equation}
where $A(v)$ is the set of all summary words aligned with $v$, and $\boldsymbol{r}_{t_v}$ and $\boldsymbol{r}_{g_v}$ are the word and node representations, respectively. Edge representations are derived for each AMR edge $(u, v) \in \mathcal{E}: \boldsymbol{r}_e = [\boldsymbol{r}_u ; \boldsymbol{r}_v]$. The edge representation $\boldsymbol{r}_e$ is fed into a classification layer that outputs a probability distribution over the output labels ($y_e$ = \{Factual, Non-Factual\}).
We assign the label \emph{non-factual} for an edge in $\mathcal{G}_{s}$ if one of the nodes in this edge is aligned with a word that belongs to a span annotated as \emph{non-factual}. Otherwise, the edge is assigned the label \emph{factual}. We call this variant \factgraphe. 

 \begin{table}[t]
\small
\centering
\resizebox{\columnwidth}{!}{
\renewcommand{\arraystretch}{0.7}
\begin{tabular}{lrc}
\toprule
\textbf{Source} & \textbf{\# datapoints} & \textbf{Domain}\\
\midrule
\citet{wang-etal-2020-asking} & 953 & CNN/DM, XSum\\
\citet{kryscinski-etal-2020-evaluating} & 1,434 & CNN/DM\\
\citet{maynez-etal-2020-faithfulness} & 2,500 & XSum\\
\citet{pagnoni-etal-2021-understanding} & 4,942 & CNN/DM, XSum\\
\midrule
Total & 9,829 & CNN/DM, XSum\\
\bottomrule
\end{tabular}}
\caption{Consolidated human annotations.} 
\label{table:statsdata}
\end{table}

\section{Experimental Setup}

%\paragraph{Data.} 
\subsection{Data} 
One of the main challenges in developing models for factuality evaluation is the lack of training data. Existing synthetic data generation approaches are not well-suited to factuality evaluation of current summarization models and human-annotated data can improve factuality models~\cite{goyal-durrett-2021-annotating}. In order to have a more effective training signal, we gather human annotations from different sources and consolidate a factuality dataset that can be used to train \factgraph and other models. 

\begin{table*}[t]
\small
\centering
\resizebox{\linewidth}{!}{\renewcommand{\arraystretch}{0.75}
\begin{tabular}{lll@{\hspace*{0.8cm}}ll@{\hspace*{0.8cm}}ll}
\toprule
\multirow{2}{*}{\textbf{Model}} & \multicolumn{2}{c@{\hspace*{0.8cm}}}{\textbf{All data}} & \multicolumn{2}{c@{\hspace*{0.8cm}}}{\textbf{CNN/DM}} & \multicolumn{2}{c}{\textbf{XSum}}  \\ \cline{2-7}
\addlinespace[0.6mm]
%\midrule
& \textbf{BACC} & \textbf{F1}  & \textbf{BACC} & \textbf{F1}  & \textbf{BACC} & \textbf{F1}  \\
\specialrule{.4pt}{0.1pt}{2pt}
\textsc{QAGS}~\cite{wang-etal-2020-asking} & 79.8 & 79.7 & 64.2 & 76.2 & 59.3 & 85.2 \\
\textsc{QUALS}~\cite{nan-etal-2021-improving} & 78.3 & 78.5 & 60.8 & 76.2 & 57.5 & 82.2 \\
\specialrule{.4pt}{0.1pt}{2pt}
\factcc~\cite{kryscinski-etal-2020-evaluating} & 76.0 & 76.3 & 69.0 & 77.8 & 55.9 & 73.9 \\
\textsc{FactCC+} & 83.9 {\scriptsize (0.4)} & 84.2 {\scriptsize (0.4)} & 68.0 {\scriptsize (1.0)}  & 83.7 {\scriptsize (0.5)} & 58.3 {\scriptsize (2.2)}  & 84.9 {\scriptsize (1.0)}     \\
\factgraph  & 86.3 {\scriptsize (1.3)} & 86.7 {\scriptsize (1.1)}  & 73.0 {\scriptsize (2.3)} & 86.8 {\scriptsize (0.8)}  & 68.6 {\scriptsize (2.3)} & 86.6 {\scriptsize (2.0)}     \\
\factgraph \hspace{0.1pt}{\scriptsize (pretrained structural adapters)}  & 86.4 {\scriptsize (0.6)} & 86.8 {\scriptsize (0.5)}  & 74.1 {\scriptsize (1.0)} & 87.4 {\scriptsize (0.3)}  & \textbf{70.4} {\scriptsize (1.9)} & 85.9 {\scriptsize (1.4)}     \\
\factgraph \hspace{0.1pt}{\scriptsize (pretrained structural and text adapters)}  & \textbf{87.6} {\scriptsize (0.7)} & \textbf{87.8} {\scriptsize (0.7)}  & \textbf{76.0} {\scriptsize (2.8)} & \textbf{87.5} {\scriptsize (0.4)}  & 69.9 {\scriptsize (2.3)} & \textbf{88.4} {\scriptsize (1.2)}     \\
\bottomrule
\end{tabular}}
\caption{BACC and F1 scores for factuality models in the test set of {\textsc{FactCollect}}. Mean ($\pm$s.d.) over 5 seeds.}
\label{table:mainresults}
%\vspace{-3mm}
\end{table*}

The source collections of the dataset are presented in Table~\ref{table:statsdata}. The dataset covers two parts, namely CNN/DM~\cite{10.5555/2969239.2969428} and XSum~\cite{nallapati-etal-2016-abstractive2}. CNN/DM contains news articles from two providers, CNN and DailyMail; while XSum contains BBC articles. CNN/DM has considerably lower levels of abstraction, and the summary exhibits high overlap with the article; a typical CNN/DM summary consists of several bullet points. In XSum, the first sentence is removed from an article and used as a summary, making it highly abstractive. After we remove duplicated annotations, the total number of datapoints is 9,567, which we divide into train (8,667), dev (300) and test (600) sets. We call this dataset {\textsc{FactCollect}}.

\subsection{Method Details} 
%\subsection{Implementation details}
%\paragraph{Selecting the Document Semantic Graphs.} 

We limit the number of considered document graphs due to efficiency reasons. In particular, we compute the pairwise cosine similarity between the embeddings of each sentence $d \in D$ and the summary sentence $S$, generated by Sentence Transformers~\cite{reimers-gurevych-2019-sentence}. We thus select $k$ sentences from the source document with the highest scores to be used to generate the document semantic graphs.

%\paragraph{Method details.} 
The model weights are initialized with {\small\textsc{Electra}} (electra-base discriminator, 110M parameters, \citeauthor{Clark2020ELECTRA:}, \citeyear{Clark2020ELECTRA:}), the structural adapters are pretrained using the release 3.0 of the AMR corpus containing 55,635 gold annotated AMR graphs, and the text adapters are pretrained using synthetic generated data. The adapters' hidden dimension is 32, which corresponds to about 1.4\% of the parameters of the original {\small\textsc{Electra}} encoders. The number of considered document graphs ($k$) is 5.\footnote{Hyperparameter details and pretraining procedures are described in Appendix~\ref{appe:hyperparameters}.} We report the test results when the balanced accuracy (BACC) on dev set is optimal. Following previous work~\cite{kryscinski-etal-2020-evaluating,goyal-durrett-2021-annotating}, we evaluate our models using BACC and Micro F1 scores. 
\vspace{-1mm}
\section{Results and Analysis}
We compare \factgraph with different methods for factuality evaluation: two QA-based methods, namely {\textsc{QAGS}} \cite{wang-etal-2020-asking} and {\textsc{QUALS}} \cite{nan-etal-2021-improving}, and \factcc \cite{kryscinski-etal-2020-evaluating}. We fine-tune \factcc using the training set, that is, it is trained on both synthetic data and {\textsc{FactCollect}}. We call this approach {\textsc{FactCC+}}. 

Table~\ref{table:mainresults} presents the results. QA-based approaches perform comparatively worse than \factcc on CNN/DM, while {\textsc{QAGS}} has a general better performance than {\textsc{QUALS}}. \factcc has a strong performance on CNN/DM, as it was trained on synthetic data derived from this dataset. However, the \factcc's performance does not transfer to XSum. {\textsc{FactCC+}} has a large increase in performance, especially on XSum, demonstrating the importance of human-annotated data for training improved factuality models.

\begin{table*}
\small
\centering
\resizebox{\linewidth}{!}{\renewcommand{\arraystretch}{0.5}
\begin{tabular}{l @{\hspace*{0.5cm}} c c c c @{\hspace*{1cm}} c c c c @{\hspace*{1cm}} c c c c }
\toprule
  & \multicolumn{4}{c@{\hspace*{1cm}}}{\textbf{All data}} & \multicolumn{4}{c@{\hspace*{1cm}}}{\textbf{CNN/DM}} & \multicolumn{4}{c}{\textbf{XSum}} \\
\midrule
\textbf{Pearson, Spearman} &$\rho$ & p-val & $r$ & p-val &$\rho$ & p-val & $r$ & p-val &$\rho$ & p-val & $r$ & p-val  \\
\midrule
{\textsc{BLEU}} & .10  & .00 & .05 & .02 & .06 & .06 & .07 & .02 & .16 & .00 & .15 & .00 \\
{\textsc{METEOR}} & .13 & .00  & .10 & .00 & .11 & .00 & .11 & .00 & .16 & .00 & .08 & .01  \\
{\textsc{Rouge-L}} & .13 & .00 & .09 & .00 & .09 & .00 & .10 & .00 & .17 & .00 & .09 & .01  \\
{\textsc{BERTScore}} & .16 & .00 & .11 & .00 & .13 & .00 & .12 & .00 & .19 & .00 & .10 & .00\\
\midrule
\smatch$_1$ & .07 & .00 & -.01  & .62 & .07 & .02 & .03 & .26 & .09 & .01 & .07 & .05 \\
\smatch$_3$ & .11 & .00 & .10 & .00 & .15 & .00 & .14 &  .00 & .06 & .10 & .04 & .21 \\
\smatch$_5$ & .13 &  .00 & .13 & .00 & .17 & .00 & .16 & .00 & .05 & .17 & .04 & .28 \\
\smatch$_{ref}$ & .08 & .00 & .03 & .20 & .05 & .12 & .03 & .35 & .13 & .00 & .08 & .02 \\
\midrule
{\textsc{QAGS}} & .22 & .00& .23 & .00 & .34 & .00 & .27 & .00 & .07 & .05 & .06 & .09\\
{\textsc{QUALS}} & .22 & .00 & .19 & .00 & .31 & .00 & .27 & .00 & .14 & .00 & .07 & .03 \\
{\textsc{DAE}} & .17 & .00 & .20 & .00 & .27 & .00 & .22 & .00 & .03 & .38 & .33 & .00 \\
{\textsc{FactCC}} & .20 & .00 & .29 & .00 & .36 & .00 & .30 & .00 &  .06 & .07 & .19 & .00\\
\midrule
{\textsc{FactCC+}} & .32 & .00 & .38 & .00 & .40 & .00 & .28 & .00 & .24 & .00 & .16 & .00\\
\factgraph & \textbf{.35} & .00 & \textbf{.42} & .00 & \textbf{.45} & .00 & \textbf{.34} & .00 &  \textbf{.30} & .00 & \textbf{.49} & .00\\
\bottomrule
\end{tabular}
}
\caption{Partial Pearson and Spearman correlation coefficients and  p-values between human judgements and methods scores for the test split of~\citet{pagnoni-etal-2021-understanding}.}
\label{tab:correlations} 
%\vspace{-3mm}
\end{table*}

\factgraph outperforms {\textsc{FactCC+}} by 2.4 BACC points in both subsets and by 10.3 BACC in XSum, even though {\textsc{FactCC+}} was pretrained on millions of synthetic examples. This indicates that considering semantic representations is beneficial for factuality evaluation and \factgraph can be trained on a small number of annotated examples. Pretraining structural adapters improves the performance on CNN/DM and XSum. Finally, \factgraph's performance further improves when both structural and text adapters are pretrained, improving over {\textsc{FactCC+}} by 3.7 BACC points.\footnote{\factgraph is significantly better than {\textsc{FactCC+}} with $p{<}0.05$ on both BACC and F1 scores.}
\vspace{-1mm}
\subsection{Correlation with Human Judgments}

We also evaluate the model performance using correlations with human judgments of factuality~\cite{pagnoni-etal-2021-understanding}. In this experiment, {\textsc{FactCC+}} and \factgraph are trained with the {\textsc{FactCollect}} data without the \citet{pagnoni-etal-2021-understanding}'s subset, which is used as dev and test sets, according to its split. For both models, following~\citet{pagnoni-etal-2021-understanding}, we obtain a binary factuality label for each sentence and take the average of these labels as the final summary score. We use the official script to calculate the correlations.\footnote{\url{https://github.com/artidoro/frank}}

\paragraph{AMR and Factuality.} We investigate whether {\textsc{Smatch}}~\cite{cai-knight-2013-smatch}, a metric that measures the degree of overlap between two AMRs, correlates with factuality judgments. We calculate the {\textsc{Smatch}} score between all the summary sentence graphs and $k$ document sentence graphs, with $k \in \{ 1,3,5 \}$. We obtain one score per summary sentence by maxing over its scores with the sentence graphs, then averaging over the summary sentence scores to obtain the summary-level score. We also calculate the {\textsc{Smatch}} between the generated summary and the reference summary graphs. As shown in Table~\ref{tab:correlations}, {\textsc{Smatch}} approaches have a small but consistent correlation, slightly improving over n-gram based metrics (e.g., {\textsc{METEOR}} and {\textsc{Rouge-L}}) in CNN/DM, suggesting that AMR, which has a higher level of abstraction than plain text, may be a semantic representation alternative to content verification.

QA-based approaches have higher correlation on the CNN/DM dataset than XSum where their correlation is relatively reduced, and {\textsc{DAE}} shows higher Spearman correlation than {\textsc{FactCC}} on XSum.
{\textsc{FactCC+}} and \factgraph, which are trained on data from {\textsc{FactCollect}}, have a overall higher performance than models trained on synthetic data, such as {\textsc{FactCC}}, again demonstrating the importance of the human-annotation signal when training factuality evaluation approaches. Finally, \factgraph has the highest correlations in both datasets, with a large improvement in XSum, suggesting that representing facts as semantic graphs is effective for more abstractive summaries.

 \begin{figure}[t]
    \centering
    \includegraphics[width=0.45\textwidth]{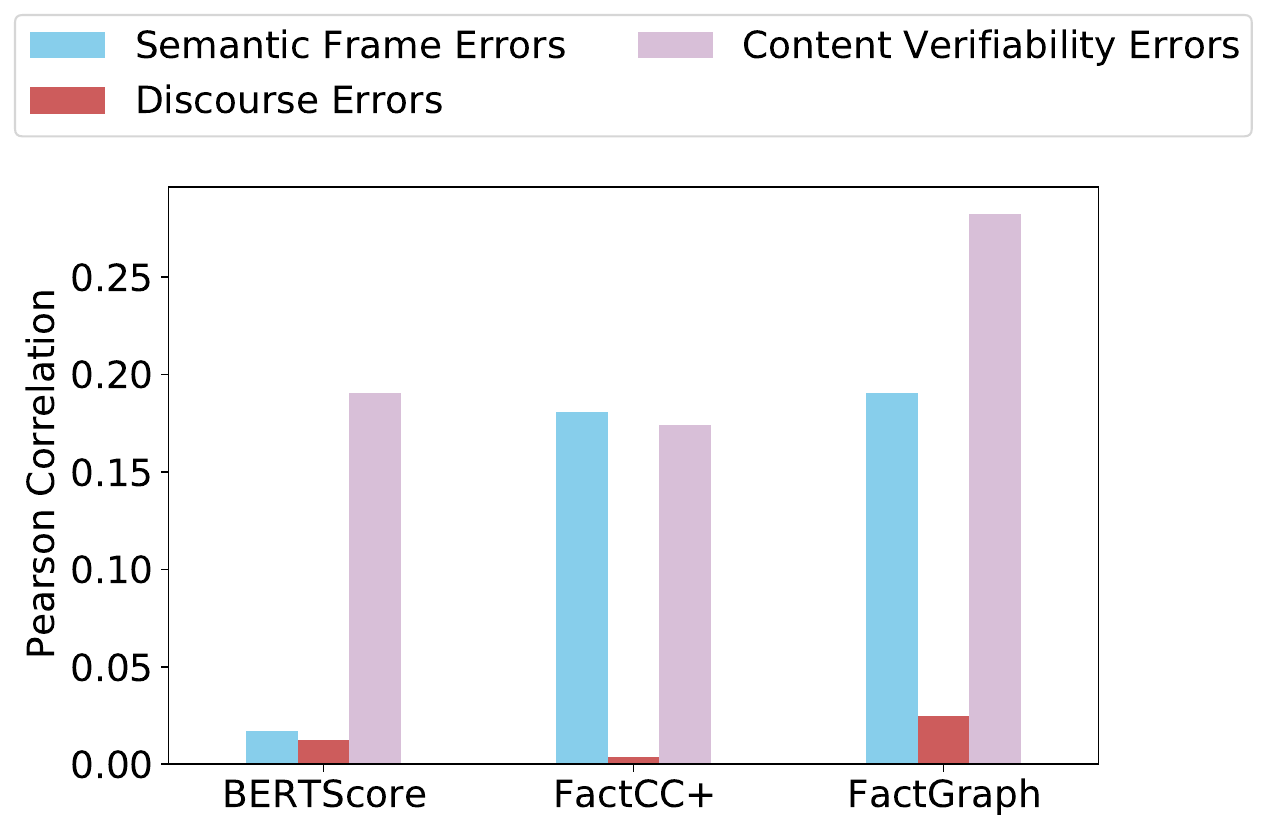}
    \caption{Variation in partial Pearson correlation when omitting error types. Higher variation indicates greater influence of an error type in the overall correlation.}
    \label{fig:factualerrors}
    %\vspace{-3mm}
\end{figure}

\paragraph{Types of Errors.} Figure~\ref{fig:factualerrors} shows the influence of the different types of factuality errors~\cite{pagnoni-etal-2021-understanding} for each approach.  \emph{Semantic Frame Errors} are errors in a frame, core, and non-core frame elements.\footnote{A semantic frame is a representation of an event, relation, or state~\cite{10.3115/980845.980860}.} \emph{Discourse Errors} extend beyond a single semantic frame introducing erroneous links between discourse segments. \emph{Content Verifiability Errors} capture cases when it is not possible to verify the summary against the source document due to the difficulty in aligning it to the source.\footnote{Refer to~\citet{pagnoni-etal-2021-understanding} for a detailed description of the error categories and the correlation computations.} Note that whereas {\textsc{BERTScore}} strongly correlates with content verifiability errors as it is a token-level similarity metric, the other methods improve in \emph{Semantic Frame Errors}. \factgraph has the highest performance suggesting that graph-based MRs are able to capture different semantic errors well. In particular, \factgraph improves in capturing content verifiability errors by 48.2\%, suggesting that representing facts using AMR is helpful.

\begin{table}[t]
\small
\centering
\begin{tabular}{lcc}
\toprule
\addlinespace[1mm]
\footnotesize\textbf{Sentence-level models} & \textbf{BACC}\\
\addlinespace[0.3mm]
Sent-Factuality~\cite{goyal-durrett-2021-annotating} & 65.6  \\
\factgraph & 74.9 \\
\midrule
\addlinespace[1mm]
\footnotesize\textbf{Edge-level models} & \textbf{BACC} \\
\addlinespace[0.3mm]
{\textsc{DAE}}~\cite{goyal-durrett-2021-annotating} & 78.7 \\
\factgraphe  & \textbf{81.1}\\

\bottomrule
\end{tabular}
\caption{Sentence-level BACC in human-annotated XSum generated summaries~\cite{maynez-etal-2020-faithfulness}. } 
\label{table:xsumannotated}
\end{table}

\subsection{Edge-level Factuality Classification}
\label{sec:results-edges}

We assess factuality beyond sentence-level with \factgraphe (\S\ref{sec:method-edges}). We train and evaluate the model against the sentence-level factuality data from~\citet{maynez-etal-2020-faithfulness}. In this dataset, human annotations for sentence and span levels are available. We derive the edge labels required for \factgraphe training as follows: For each edge in the summary graph, if one of the nodes connected to this edge is aligned with a word that belongs to a span labeled as non-factual, the edge is annotated as non-factual.\footnote{We use the JAMR aligner~\cite{flanigan-etal-2014-discriminative} to obtain node-to-word alignments.} Summary-level labels are obtained from edge-level predictions: if any edge in the summary graph is classified as non-factual, the summary is labeled as non-factual. We use the same splits from~\citet{goyal-durrett-2021-annotating}.\footnote{We sample 100 datapoints from the training set as dev set to execute hyperparameter search.} We compare \factgraphe with {\textsc{DAE}} and additionally with a sentence-level baseline~\cite{goyal-durrett-2021-annotating} and \factgraph.

Table~\ref{table:xsumannotated} shows that the edge-level factuality classification gives better performance than sentence-level classification, and \factgraph performs better in both sentence and edge classification levels. \factgraphe outperforms {\textsc{DAE}}, demonstrating that training on subsentence-level factuality annotations enables it to accurately predict edge-level factuality and output summary-level factuality. 

Finally, while the semantic representations contribute to overall performance, extracting those representations adds some overhead in preprocessing time (and slightly more in inference time), as shown in Appendix~\ref{appe:speedcomparison}.

\subsection{Model Ablations}
\label{sec:model_eval}

In Table~\ref{table:ablation}, we report an ablation study on the impact of distinct \factgraph's components. First, note that only encoding the textual information leads to better performance than just encoding graphs. This is expected since pretrained encoders are known for good performance in NLP textual tasks due to their transfer learning capabilities and the full document text encodes more information than the selected $k$ document graphs. Moreover, AMR representations abstract aspects such as verb tenses, making the graphs agnostic regarding more fine-grained information. However, this is compensated in \factgraph, which captures coarse-grained details from the text modality. Future work can consider incorporating such information into the graph representation in order to improve the factuality assessment. 

\begin{table}[t]
%\vspace{-1mm}
\small
\centering
\begin{tabular}{lcc}
\toprule
\textbf{Model} & \textbf{BACC} & \textbf{F1} \\
\midrule
Only graphs & 77.7 & 78.0 \\
Only text & 88.4 & 88.6 \\
 \factgraph& 91.2 & 91.3 \\
\bottomrule
\end{tabular}
\vspace{-1mm}
\caption{Ablation study for different components of the model in the {\textsc{FactCollect}}'s dev set.} 
\label{table:ablation}
\vspace{-4mm}
\end{table}

Ultimately, \factgraph, which uses both document and summary graphs, gives the overall best performance, demonstrating that semantic graph representations complement the text representation and are beneficial for factuality evaluation.

 \begin{figure*}[t]
    \centering
    \includegraphics[width=1\textwidth]{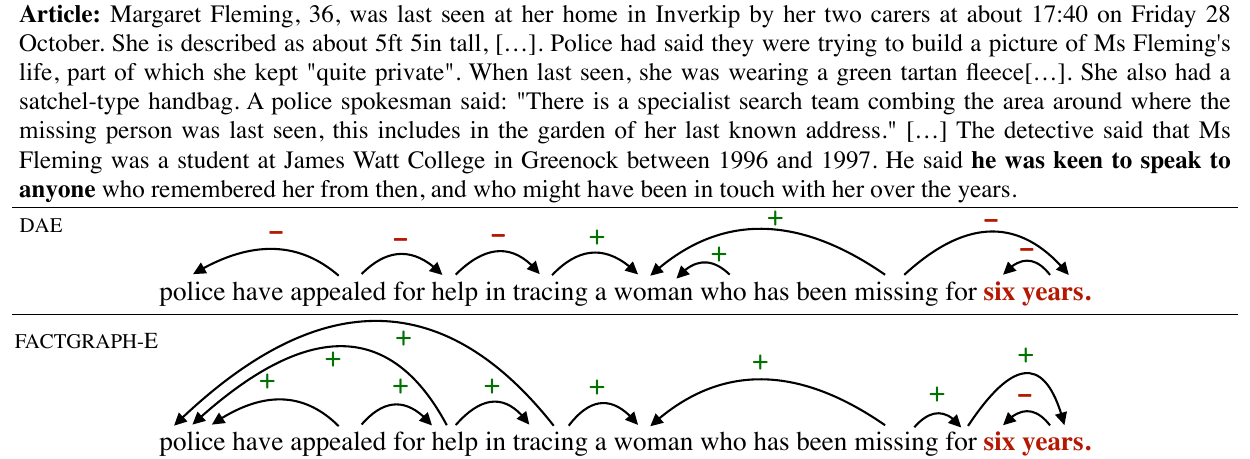}
    \caption{An example of a document, its generated summary and factuality predictions for word pairs, based on the dependency graph ({\textsc{DAE}}) versus AMR graph (\factgraphe). $+/-$ means the predicted label for that edge. 
    }
    \label{fig:casestudy}
    \vspace{-1mm}
\end{figure*}

\paragraph{Number of Document Graphs}
\label{appe:docgraph}
Table~\ref{table:effect-num-graphs} shows the influence of the number of considered document graphs measured on {\textsc{FactCollect}}'s dev set performance. Note that generally more document graphs leads to better performance with a peak in 5. This suggests that using all graph sentences from the source document is not required for better performance. Moreover, the results indicate that our strategy of selecting document graphs using the contextual representations of the document sentences which are compared to the summary performs well in practice.

We additionally present the performance of \factgraph with other semantic representations in Appendix~\ref{appe:semanticreps}.

\subsection{Comparison to Full Fine-tuning}
\label{appe:compfinetuning}
\factgraph only trains adapter weights that are placed into each layer of both text and graph encoders. We compare \factgraph with a model with similar architecture, with both text and graph encoders, but without (structural) adapter layers. We then fine-tune all the model parameters. Table~\ref{tab:finetuned} shows that \factgraph performs better even though it trains only 1.4\% of the parameters of the fully fine-tuned model, suggesting that the structural adapters help to adapt the graph encoder to semantic graph representations.

\begin{table}[t]
\small
\centering
\begin{tabular}{lcc}
\toprule
\textbf{\# Graphs} & \textbf{BACC} & \textbf{F1}\\
\midrule
1  & 90.1 & 90.3 \\
3 &	90.9 & 91.0  \\
5 (\textbf{final})& 91.2 & 91.3  \\
7 &	89.8 & 90.0  \\
\bottomrule
\end{tabular}
\caption{Effect in the {\textsc{FactCollect}}'s dev set of the number of considered AMR graphs from the document.} 
\label{table:effect-num-graphs}
\vspace{-3mm}
\end{table}

\subsection{Case Study}
\label{sec:casestudy}
\factgraphe computes factuality scores for each edge of the AMR summary graph and those predictions are aggregated to generate a sentence-level label (\S\ref{sec:results-edges}). Alternatively, it is possible to identify specific inconsistencies in the generated summary based on the AMR graph structure. This factuality information at subsentence-level can provide deeper insights on the kinds of factual inconsistencies made by different summarization models~\cite{maynez-etal-2020-faithfulness} and can supply text generation approaches with localized signals for training~\cite{cao-etal-2020-factual,zhu-etal-2021-enhancing}.

Figure~\ref{fig:casestudy} shows a document, its generated summary, and factuality edge predictions by {\textsc{DAE}} and \factgraphe.\footnote{Appendix~\ref{appe:semanticrepsamrdep} presents the complete AMR and dependency summary graphs.} First, note that since {\textsc{DAE}} uses dependency arcs and \factgraphe is based on AMR, the sets of edges in both approaches, that is, the relations between nodes and hence words, are different. Second, both methods are able to detect the hallucination \emph{six years}, which was never mentioned in the source document. However, {\textsc{DAE}} does not consider that \emph{police appealed for help in tracing} is factual whereas \factgraphe captures it. This piece of information is related to a span in the document with a very different but semantically related form (highlighted in bold in Figure~\ref{fig:casestudy}). This poses challenges to {\textsc{DAE}}, since it classifies semantic relations independently and only considers the text surface. On the other hand, \factgraphe matches the summary against the document not only at text surface level but semantic level. 

\begin{table}[t]
\small
\centering
\begin{tabular}{lccr}
\toprule
         & \textbf{BACC} & \textbf{F1} & \textbf{Parameters}\\
          \midrule
          Fully fine-tuned & 90.3 & 90.3 & 100.0\%\\
           \factgraph & 91.2 & 91.3 & 1.4\%\\
\bottomrule
\end{tabular}
\caption{Comparison between \factgraph and fully fine-tuning in the dev set of {\textsc{FactCollect}}.}
\label{tab:finetuned}
\vspace{-3mm}
\end{table}

\section{Conclusion}

We presented \factgraph, a graph-based approach to explicitly encode facts using meaning representations to identify factual errors in generated text. We provided an extensive evaluation of our approach and showed that it significantly improves results on different factuality benchmarks for summarization, indicating that structured semantic representations are beneficial to factuality evaluation. Future work includes (i) exploring approaches to develop document-level semantic graphs~\cite{naseem2021docamr}, (ii) an explainable graph-based component to highlight hallucinations and (iii) to combine different meaning representations in order to capture distinct semantic aspects.

\section*{Acknowledgments}
We thank the anonymous reviewers for their valuable feedback. We also thank Shiyue Zhang, Kevin Small, and Yang Liu for their feedback on this work. Leonardo F. R. Ribeiro has been supported by the German Research Foundation (DFG) as part of the Research Training Group ``Adaptive Preparation of Information form Heterogeneous Sources'' (AIPHES, GRK 1994/1). 

\section*{Impact Statement}

In this paper, we study the problem of detecting factual inconsistencies in summaries generated from input documents. The proposed models better consider the text internal meaning structure and could benefit general generation applications by evaluating their output regarding factual consistency, which could ensure that these systems are more trustworthy. This work is built using semantic representations extracted using AMR parsers. In this way, the quality of the parser used to generate the semantic representations can significantly impact the results of our models. In our work, we mitigate this risk by employing a state-of-the-art AMR parser~\cite{Micheleamr}.

\bibliography{anthology,custom}
\bibliographystyle{acl_natbib}

%\clearpage
\vspace{10pt}
\appendix

\section*{Appendices}

In this supplementary material, we detail experiments' settings, additional model evaluations and additional information about semantic graph representations.

\section{Additional Examples}
\label{appe:examples_amrs}
\begin{figure*}[t]
    \centering
    \includegraphics[width=1\textwidth]{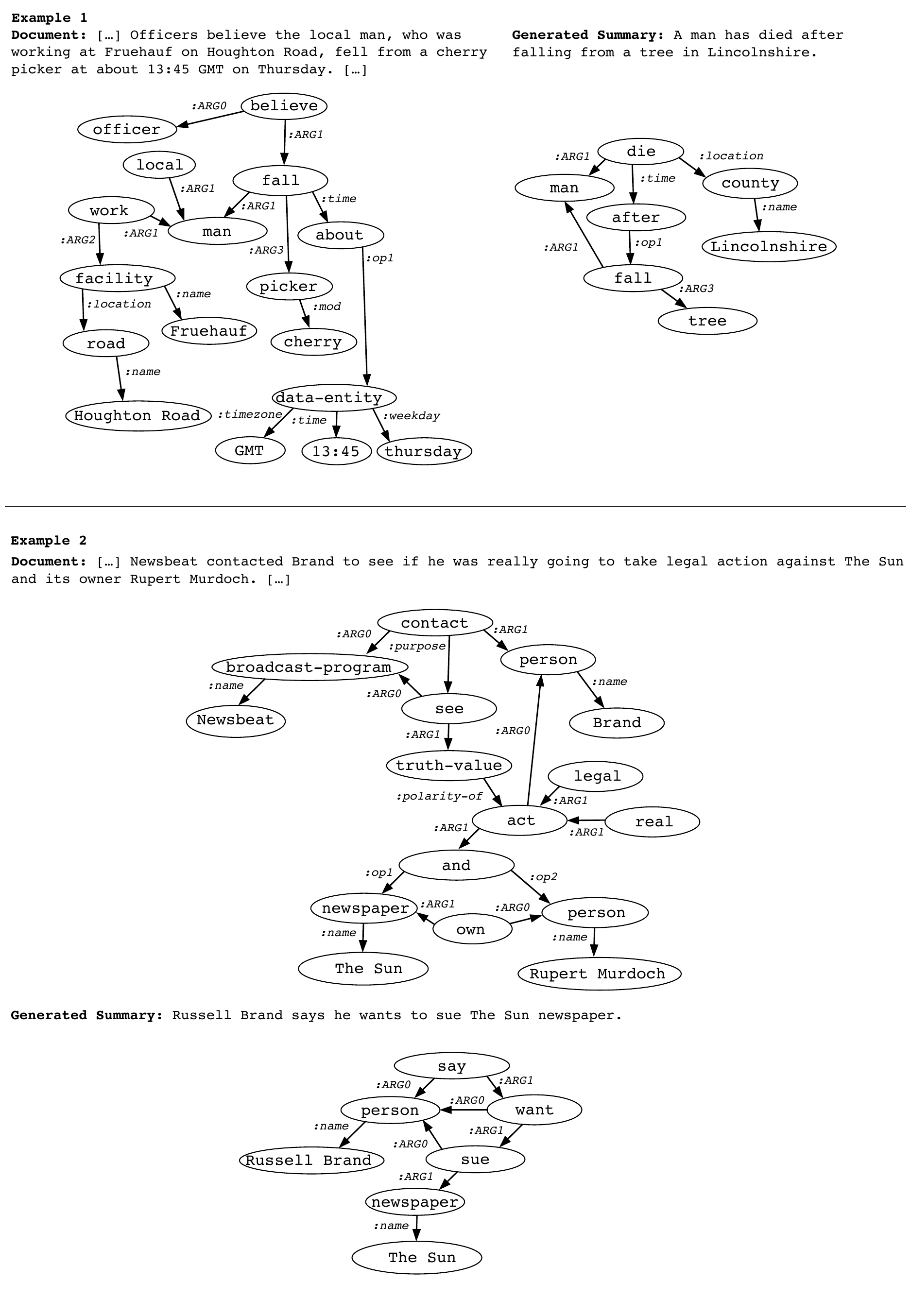}
    \caption{Examples of graph-based meaning representations parsed from sentences of documents and generated summaries.}
    \label{fig:examples_appendix}
\end{figure*}

Figure~\ref{fig:examples_appendix} shows examples of AMR representations generated from summaries and salient sentences from the respective source document.

\section{Details of Models and Hyperparameters} % (fold)
\label{appe:hyperparameters}
The experiments were executed using the version $3.3.1$ of the \emph{transformers} library released by Hugging Face \citep{wolf2019huggingfaces}. In Table \ref{tab:hyper}, we report the hyperparameters used to train \factgraph. We use the Adam optimizer \cite{kingma:adam} and employ a linearly decreasing learning rate schedule without warm-up. Mean pooling is used to calculate the final representation of each graph.

\begin{table}[t]
\small
\centering
\resizebox{\linewidth}{!}{
\begin{tabular}{lc}
\toprule
\textbf{Computing Infrastructure} & 32GB NVIDIA V100 GPU \\
\textbf{Optimizer} & Adam\\
\textbf{Optimizer Params} & $\beta=(0.9, 0.999), \epsilon=10^{-8}$ \\
          \textbf{learning rate} & 1e-4\\
          \textbf{Learning Rate Decay} & Linear \\
                  \textbf{Weight Decay} & 0 \\
        \textbf{Warmup Steps} & 0 \\
        \textbf{Maximum Gradient Norm} & 1 \\
          \textbf{batch size} & 4\\
           \textbf{epoch} & 10\\
           \textbf{Adapter dimension} & 32\\
           \textbf{\# document graphs ($k$)} & 5\\
\bottomrule
\end{tabular}}
\caption{Hyperparameter settings for our methods. }
\label{tab:hyper}

\end{table}

\paragraph{Structural Adapters' Pretraining.} The structural adapters are pretrained using AMR graphs from the release 3.0 (LDC2020T02) of the AMR annotation corpus~\cite{amr3}.\footnote{\href{https://catalog.ldc.upenn.edu/LDC2020T02}{https://catalog.ldc.upenn.edu/LDC2020T02}} Similarly to the masked language modeling objective, we execute self-supervised node-level prediction, where we randomly mask and classify AMR nodes. The goal of this pretraining phase is to capture domain specific AMR knowledge by learning the regularities of the node/edge attributes distributed over graph structure.

\paragraph{Text Adapters' Pretraining.} The text adapters are pretrained using synthetically created data, which is generated by applying a series of rule-based transformations to the sentences of source documents~\cite{kryscinski-etal-2020-evaluating}. The pretraining task is to classify each summary sentence as factual or non-factual. The goal of this pretraining phase is to learn suitable text representations to better identify whether summary sentences remain factually consistent to the input document after the transformation.

\section{Speed Comparison}
\label{appe:speedcomparison}

\factgraph encodes the structured semantic representations that encode facts from the document and summary. Despite their effectiveness, extracting semantic graphs, such as AMR, is computationally expensive because current models employ Transformer-based encoder-decoder architectures based on Transformers and pretrained language models. 

In this experiment, we compare the time execution of \factgraphe and {\textsc{DAE}} in a sample of 1000 datapoints extracted from the XSum test set. In order to extract the semantic graphs, we investigate two AMR parsers, \textbf{Parser1}: a dual graph-sequence parser that iteratively refines an incrementally constructed graph \cite{cai-lam-2020-amr} , and \textbf{Parser2}: a linearized graph model that employs BART~\cite{Micheleamr}. The execution of the AMR parsers is parallelized using four Tesla V100 GPUs. We use Parser2 for the experiments in this paper since it is the current state of the art in AMR parsing, although it is slower in preprocessing than Parser1.

\begin{table}[t]
\small
\centering
\resizebox{\linewidth}{!}{
\begin{tabular}{lrr}
\toprule
          &  \textbf{Preprocessing} &  \textbf{Inference}\\
          \midrule
          {\textsc{DAE}} & 135.8 & 62.6\\
           \factgraphe \,- \textbf{Parser1}  & 427.9 & 79.0\\
           \factgraphe \,- \textbf{Parser2}  & 1332.2 & 75.4\\
\bottomrule
\end{tabular}}
\caption{Speed Comparison. Execution time is measured in seconds. }
\label{tab:speendcomparison}
\end{table}

\begin{table}[t]
\small
\centering
\renewcommand{\arraystretch}{0.9}
\begin{tabular}{lcc}
\toprule
\textbf{Graph Type} & \textbf{BACC} & \textbf{F1} \\
%\midrule
\midrule
Only text  & 88.4 & 88.6 \\
\factgraph-Dependency & 90.2 & 90.3  \\
\factgraph-OpenIE & 90.5 & 90.7\\
\factgraph-AMR &  91.2 & 91.3 \\
\bottomrule
\end{tabular}
\caption{Effect of different graph representations in the factuality model (on the dev set of {\textsc{FactCollect}}).} 
\label{table:effect-graph-rep}
\end{table}

As shown in Table~\ref{tab:speendcomparison}, {\textsc{DAE}}'s preprocessing is much faster compared to this phase in \factgraphe, since {\textsc{DAE}} employs a fast enhanced dependency model from the Stanford CoreNLP tool~\cite{manning-etal-2014-stanford}. This model builds a parse by performing a linear-time scan over the words of a sentence. Finally, note that \factgraph is slower than {\textsc{DAE}} in inference because it employs adapters and encodes both graphs and texts from the document and summary, whereas the {\textsc{DAE}} model encodes only the texts. 

\section{Comparing Semantic Representations for Factuality Evaluation}
\label{appe:semanticreps}
OpenIE graph-based structures were used in order to improve factuality in abstractive summarization~\cite{Cao_Wei_Li_Li_2018}, whereas dependency arcs were shown to be beneficial for evaluating factuality~\cite{goyal-durrett-2020-evaluating}. We thus investigate different graph-based meaning representations using \factgraph. AMR is a more logical representation that models relations between core concepts, and has a rough alignment between nodes and spans in the text. Conversely, dependencies capture more fine-grained relations between words, and all words are mapped into nodes in the dependency graph. OpenIE constructs a graph with node descriptions similar to the original text and uses open-domain relations, leading to relations that are hard to compare.

As shown in Table~\ref{table:effect-graph-rep}, 
whereas OpenIE performs slightly better than dependency graphs, AMR gives the best results according to the two metrics, highlighting the potential use of AMRs in representing salient pieces of information. Different from our work, \citet{lee2021analysis} and \citet{naseem2021docamr} propose a graph construction approach which generates a single document-level graph created using the individual sentences’ AMR graphs by merging identical concepts – this is orthogonal to our sentence-level AMR representation and can be incorporated in future work. 

 \begin{figure}[t]
    \centering
    \includegraphics[width=0.49\textwidth]{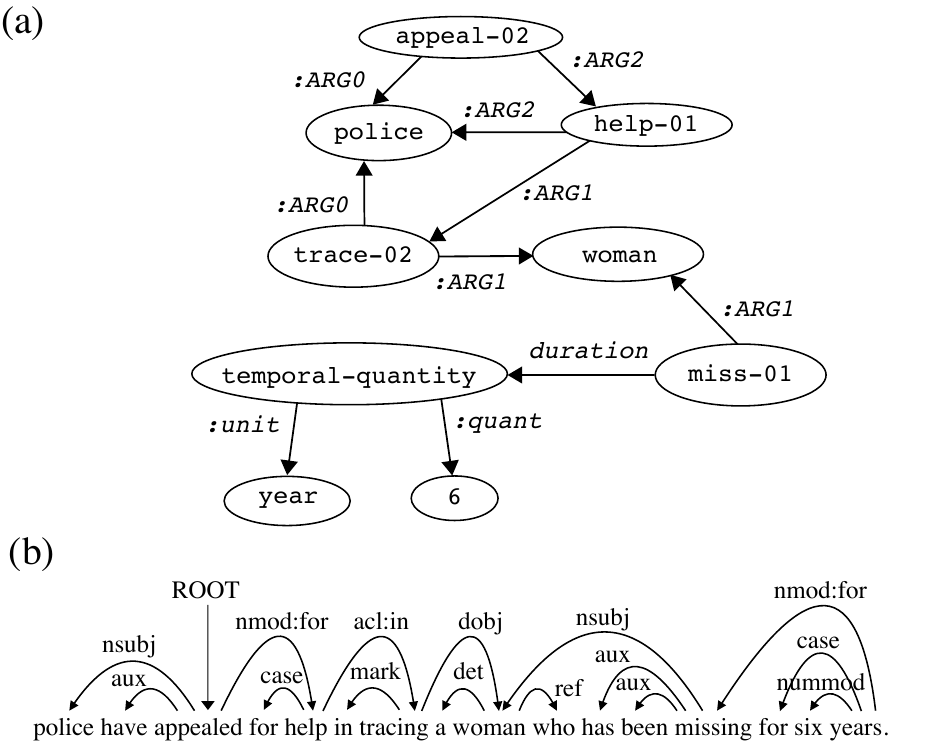}
    \caption{(a) AMR and (b) dependency representations for the summary \textit{``police have appealed for help in tracing a woman who has been missing for six years.''}
    } 
    \label{fig:casestudy-graphs}
\end{figure}

\section{Semantic Representations}
\label{appe:semanticrepsamrdep}
In Figure~\ref{fig:casestudy-graphs} we show AMR and dependency representations for the summary sentence \textit{``police have appealed for help in tracing a woman who has been missing for six years.''}. In \S\ref{sec:casestudy} those semantic representations are used to predict subsentence-level factuality using edge-level information. In particular, \factgraphe employs AMR (Figure~\ref{fig:casestudy-graphs}a) whereas {\textsc{DAE}} uses dependencies (Figure~\ref{fig:casestudy-graphs}b).

\end{document}